\documentclass{article}

\usepackage[dblblindworkshop, nonanonymous]{neurips_2026}

\usepackage[utf8]{inputenc} 
\usepackage[T1]{fontenc}    
\usepackage{hyperref}       
\usepackage{url}            
\usepackage{booktabs}       
\usepackage{amsfonts}       
\usepackage{nicefrac}       
\usepackage{microtype}      
\usepackage{xcolor}         
\usepackage{graphicx}

\nolinenumbers            

\title{ASIRF: An Agentic Framework for Context-Dependent Sensitive Information Redaction}

\author{%
  Sudha Priyadarshini\textsuperscript{1} \qquad
  Mohamed Chahine Ghanem\textsuperscript{2,1} \\[4pt]
  \textsuperscript{1}University of Liverpool, UK \qquad
  \textsuperscript{2}Keele University, UK \\
}

\begin{document}

\maketitle

\begin{abstract}
Sensitive information is defined by domain and intent, not a universal category, yet redaction systems such as privacy filters and named-entity recognizers fix a taxonomy at training time, requiring retraining for each new domain. We introduce ASIRF (Agentic Sensitive Information Redaction Framework), which retrieves domain-specific definitions based on the input's domain from a flexible knowledge base at inference time, needing no retraining to adapt. Two architectures, a three-call multi-agent pipeline and a single-agent variant, are evaluated across ten small open-weight models and eight datasets, including out-of-distribution fictional domains, against the OpenAI Privacy Filter (OPF) as a trained-classifier baseline. With only a few dozen expert-authored definitions per domain and no training data, ASIRF's recall exceeds OPF's in 68 of 80 model-domain combinations (85 percent), by at least one of the two architectures, with shortfalls confined mostly to OPF's training-distribution domains.

\end{abstract}

\section{Introduction}

Sensitive information is not an intrinsic property of a text but one assigned by domain and intent.
A ten-digit number is a phone number in a support transcript or an account identifier in a banking record.
The definitions of sensitivity change over time and across jurisdictions and domains. Trained detectors fix a taxonomy in their weights at training time, and extending it to a new domain or attribute type requires new labeled data and an expensive retraining pass. ASIRF, the Agentic Sensitive Information Redaction Framework, relocates that taxonomy to an inference-time knowledge base, instantiating a claim that adapting to a new domain requires only a knowledge-base edit, using definitions a domain expert can produce, without labeled examples or a training run.

The evaluation of ASIRF is organized around two research questions:

\begin{itemize}
  \item \textbf{RQ1.} How does an test-time agentic framework, combining an agentic
    harness with domain-conditioned retrieval, improve domain-specific
    filtering of sensitive information over conventional, training-centric
    approaches?
  \item \textbf{RQ2.} To what extent does an agentic framework generalize to
    out-of-distribution domains and previously unseen sensitive-attribute
    definitions?
\end{itemize}

RQ1 is answered in \S\ref{sec:asirfvsopf} by comparing both of ASIRF's
architectures (multi-agent and single-agent) against the OpenAI Privacy
Filter (OPF)~\cite{de2026model}, a model trained on the
identity-attribute vocabulary, and by the
retrieval-removal ablation (\S\ref{sec:ablation}). RQ2 is addressed by
the same comparison, on two out-of-distribution and fictional domains.

\section{Related Work}

Trained sensitive-information detectors pair a fixed, hand-enumerated entity list with statistical or lightweight neural classifiers, as in OPF~\cite{de2026model} and Presidio-style recognizers~\cite{presidio}, or learn a fixed label schema, as in named-entity recognizers such as GLiNER~\cite{zaratiana-etal-2024-gliner} or clinical taggers~\cite{kim2024generalizing}.
Every case requires new labeled data and retraining to add a category, none accepting one at test-time.
Retrieval-augmented generation instead grounds a language model's output in documents fetched at inference time~\cite{lewis2020retrieval}.
ASIRF applies this to entity definitions, making the sensitivity taxonomy itself an inference-time input.

Decomposing a task into staged, tool-using agent calls improves
multi-step reasoning~\cite{yao2022react}~\cite{schick2023toolformer}~\cite{shinn2023reflexion}.
ASIRF-Multi instantiates this pattern, while ASIRF-Single tests whether the benefit persists once the staged structure is collapsed (\S\ref{sec:asirfvsopf}).
The primary contribution of ASIRF is the composition of retrieval and domain-conditioned routing into a single pipeline for sensitive-information filtering.

\section{Methodology}

\begin{figure}[t]
\centering
\includegraphics[width=\linewidth]{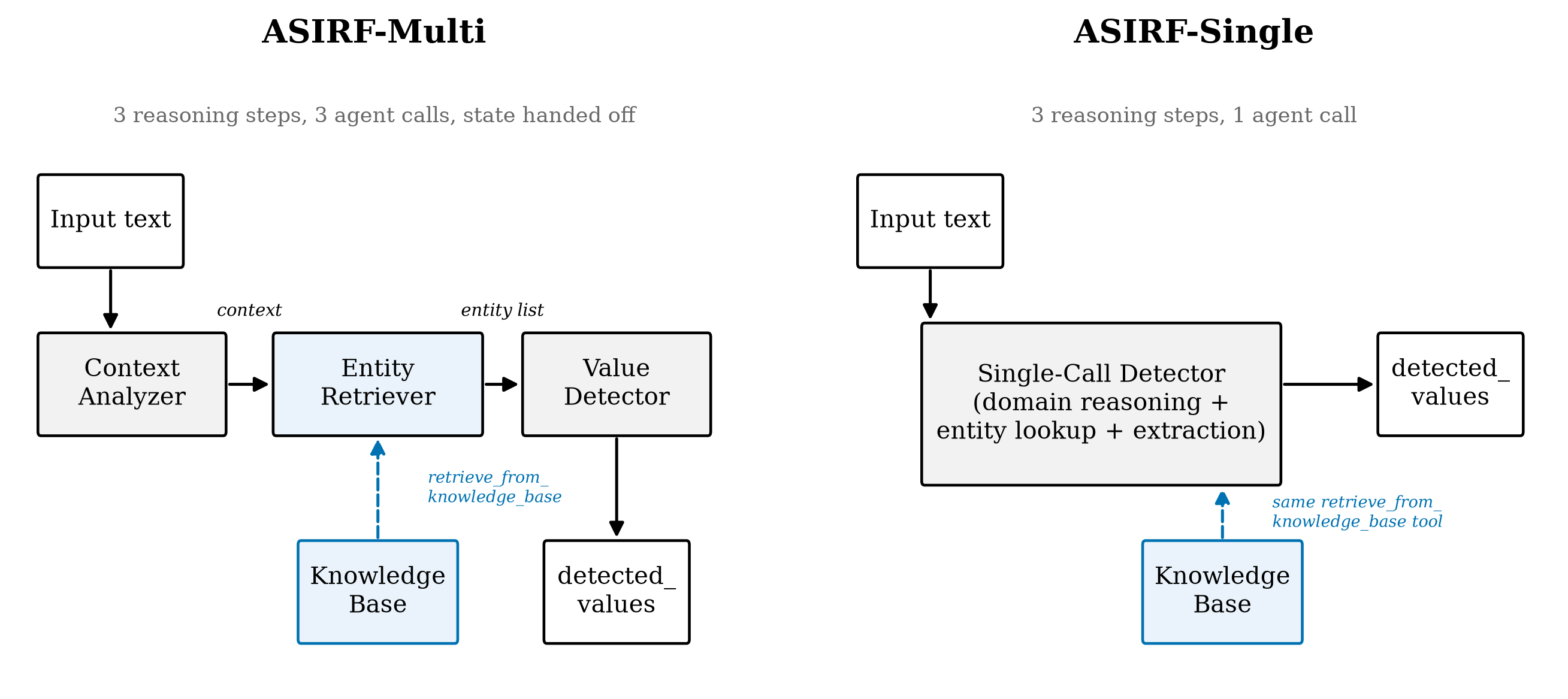}
\caption{Two architectures of ASIRF - ASIRF-Multi as three-step pipeline (classify, retrieve, extract) and ASIRF-Single as a single agent call}
\label{fig:architecture}
\end{figure}

ASIRF, as described in Figure \ref{fig:architecture}, performs three logical steps on every input, (i) classify the domain, (ii) retrieve the relevant sensitive-attribute definitions, and (iii) extract the matching values (Figure 1).
\textbf{ASIRF-Multi} implements each step as a separate chained agent (Context Analyzer, Entity Retriever, Value Detector), with only the Entity Retriever calling the retrieval tool.
\textbf{ASIRF-Single} performs all three in one call with the same tool, removing only the staged structure, not retrieval access, isolating decomposition of the steps from retrieval (\S\ref{sec:results}).
Retrieval is a semantic search over a ChromaDB~\cite{chroma2022} collection of entity definitions, indexed via an HNSW (hierarchical navigable small-world) graph~\cite{malkov2018efficient} under cosine similarity with a 1024-dimensional embeddings model~\cite{aws2024titanembeddings}, returning the top-k nearest entries de-duplicated by name.
Both architectures are evaluated across ten open-weight models spanning three families (Gemma-3~\cite{gemmateam2025gemma3}, Qwen3~\cite{qwenteam2025qwen3}, Ministral~\cite{ministral2026ministral3}, 1B–14B parameters) and two inference providers (Amazon Bedrock and Hugging Face), isolating the architecture's effect from any single model's idiosyncrasies.

The knowledge base is ASIRF's adaptation mechanism, where a new domain is onboarded by adding new entries instead of retraining.
It contains cross-domain entity types, statutory-text entries, and dataset-specific entries, including the two fictional domains (\S\ref{sec:oodgen}), under a common schema.

\section{Evaluation}
\label{sec:evaluation}

\subsection{Experimental setup}

\textbf{Baseline:} Both architectures are compared against OpenAI Privacy Filter (OPF)~\cite{de2026model}, a locally-run multi-class token classifier representing the training-centric approach that uses a fixed, pre-trained label set and activates 50M parameters in a single forward pass.
OPF runs on identical input under the same scoring rules as both ASIRF
architectures, so any difference reflects detection capability rather
than a scoring convention.

\textbf{Datasets:} Evaluation spans eight datasets.
Five are synthetically generated out of which three are modeling real-world domains (finance, healthcare, cybersecurity), and two fictional domains (Interstellar Immigration Bureau, Archive of Forgotten Futures) built with invented terminology to test out-of-distribution generalization relative to OPF~\cite{3737916.3738680}.
The remaining three, pii-masking-300k~\cite{ai4privacy_2024}, SPY~\cite{savkin-etal-2025-spy} and CredData~\cite{samsung_creddata_2021}, are established, externally sourced privacy-evaluation datasets.

\textbf{Metrics:} Precision, recall, and F1 are reported using token-level fuzzy overlap rather than exact span matching to reduce sensitivity to incidental span-boundary differences while preserving whether the sensitive value was identified~\cite{pilan-etal-2022-text}. 
Recall is foregrounded throughout, since a missed value is a severe compliance failure while a false positive is mild over-redaction.

\subsection{Results}
\label{sec:results}

\subsubsection{ASIRF vs. OPF}
\label{sec:asirfvsopf}

\begin{table}[t]
\centering
\scriptsize
\caption{ASIRF-Multi recall by model and domain. The top-performing model exceeds OPF on every dataset.}
\label{tab:multi-recall}
\resizebox{\linewidth}{!}{%
\begin{tabular}{lcccccccc}
\toprule
& \multicolumn{6}{c}{Real-world domains} & \multicolumn{2}{c}{Fictional domains} \\
\cmidrule(lr){2-7} \cmidrule(lr){8-9}
Model & Finance & Healthcare & Cybersecurity & pii-masking & SPY & CredData & \shortstack{Archive of\\Forgotten Futures} & \shortstack{Interstellar\\Immigration Bureau} \\
\midrule
ministral-14b & \textbf{0.801} & \textbf{0.761} & \textbf{0.936} & \underline{0.821} & \underline{0.922} & \textbf{0.972} & \textbf{0.908} & \textbf{0.911} \\
qwen3-14b     & 0.610 & 0.629 & 0.732 & 0.626 & 0.726 & 0.900 & 0.783 & 0.807 \\
gemma-3-12b   & \underline{0.676} & 0.581 & \underline{0.923} & \textbf{0.826} & \textbf{0.947} & 0.960 & \underline{0.902} & \underline{0.878} \\
ministral-8b  & 0.664 & 0.626 & 0.806 & 0.777 & 0.869 & 0.951 & 0.832 & 0.846 \\
qwen3-8b      & 0.621 & 0.694 & 0.787 & 0.705 & 0.698 & 0.891 & 0.754 & 0.791 \\
qwen3-4b      & 0.641 & \underline{0.716} & 0.829 & 0.787 & 0.811 & 0.942 & 0.794 & 0.845 \\
gemma-3-4b    & 0.577 & 0.449 & 0.764 & 0.750 & 0.855 & \underline{0.966} & 0.786 & 0.772 \\
ministral-3b  & 0.641 & 0.550 & 0.781 & 0.637 & 0.716 & 0.957 & 0.788 & 0.764 \\
qwen3-1.7b    & 0.577 & 0.571 & 0.739 & 0.667 & 0.635 & 0.904 & 0.668 & 0.728 \\
gemma-3-1b    & 0.462 & 0.397 & 0.357 & 0.351 & 0.130 & 0.520 & 0.471 & 0.450 \\
\midrule
\textbf{OPF}  & 0.270 & 0.206 & 0.530 & 0.800 & 0.863 & 0.861 & 0.217 & 0.348 \\
\bottomrule
\end{tabular}%
}
\end{table}

\begin{table}[t]
\centering
\scriptsize
\caption{ASIRF-Single recall by model and domain. The top-performing model exceeds OPF on every dataset.}
\label{tab:single-recall}
\resizebox{\linewidth}{!}{%
\begin{tabular}{lcccccccc}
\toprule
& \multicolumn{6}{c}{Real-world domains} & \multicolumn{2}{c}{Fictional domains} \\
\cmidrule(lr){2-7} \cmidrule(lr){8-9}
Model & Finance & Healthcare & Cybersecurity & pii-masking & SPY & CredData & \shortstack{Archive of\\Forgotten Futures} & \shortstack{Interstellar\\Immigration Bureau} \\
\midrule
ministral-14b & \textbf{0.699} & \textbf{0.671} & 0.930 & 0.791 & \textbf{0.930} & 0.972 & \textbf{0.891} & \textbf{0.905} \\
qwen3-14b     & 0.595 & 0.452 & 0.864 & 0.727 & 0.849 & 0.962 & 0.793 & 0.765 \\
gemma-3-12b   & 0.645 & 0.528 & 0.805 & 0.674 & 0.853 & \textbf{0.983} & 0.802 & 0.760 \\
ministral-8b  & 0.623 & 0.494 & 0.827 & 0.713 & 0.779 & 0.972 & 0.776 & 0.781 \\
qwen3-8b      & \underline{0.684} & \underline{0.660} & \textbf{0.943} & \underline{0.805} & \underline{0.916} & 0.979 & 0.840 & \underline{0.872} \\
qwen3-4b      & 0.674 & 0.562 & \underline{0.935} & \textbf{0.815} & 0.875 & 0.977 & \underline{0.874} & 0.858 \\
gemma-3-4b    & 0.601 & 0.476 & 0.751 & 0.581 & 0.570 & \underline{0.981} & 0.765 & 0.719 \\
ministral-3b  & 0.595 & 0.469 & 0.767 & 0.567 & 0.590 & 0.957 & 0.731 & 0.705 \\
qwen3-1.7b    & 0.668 & 0.533 & 0.888 & 0.765 & 0.834 & 0.964 & 0.815 & 0.824 \\
gemma-3-1b    & 0.657 & 0.583 & 0.732 & 0.583 & 0.405 & 0.821 & 0.802 & 0.812 \\
\midrule
\textbf{OPF}  & 0.270 & 0.206 & 0.530 & 0.800 & 0.863 & 0.861 & 0.217 & 0.348 \\
\bottomrule
\end{tabular}%
}
\end{table}

ASIRF-Multi and ASIRF-Single are each compared against OPF in Tables 1–2.

ASIRF exceeds OPF's recall in 68 of 80 model–domain combinations (85\%), by at least one of the two ASIRF architectures.
The shortfalls are concentrated on the two domains matching OPF's training data~\cite{de2026model}, pii-masking~\cite{ai4privacy_2024} and SPY datasets~\cite{savkin-etal-2025-spy}, where only the largest
models close the gap (Figure 2).

\begin{figure}[t]
\centering
\includegraphics[width=\linewidth]{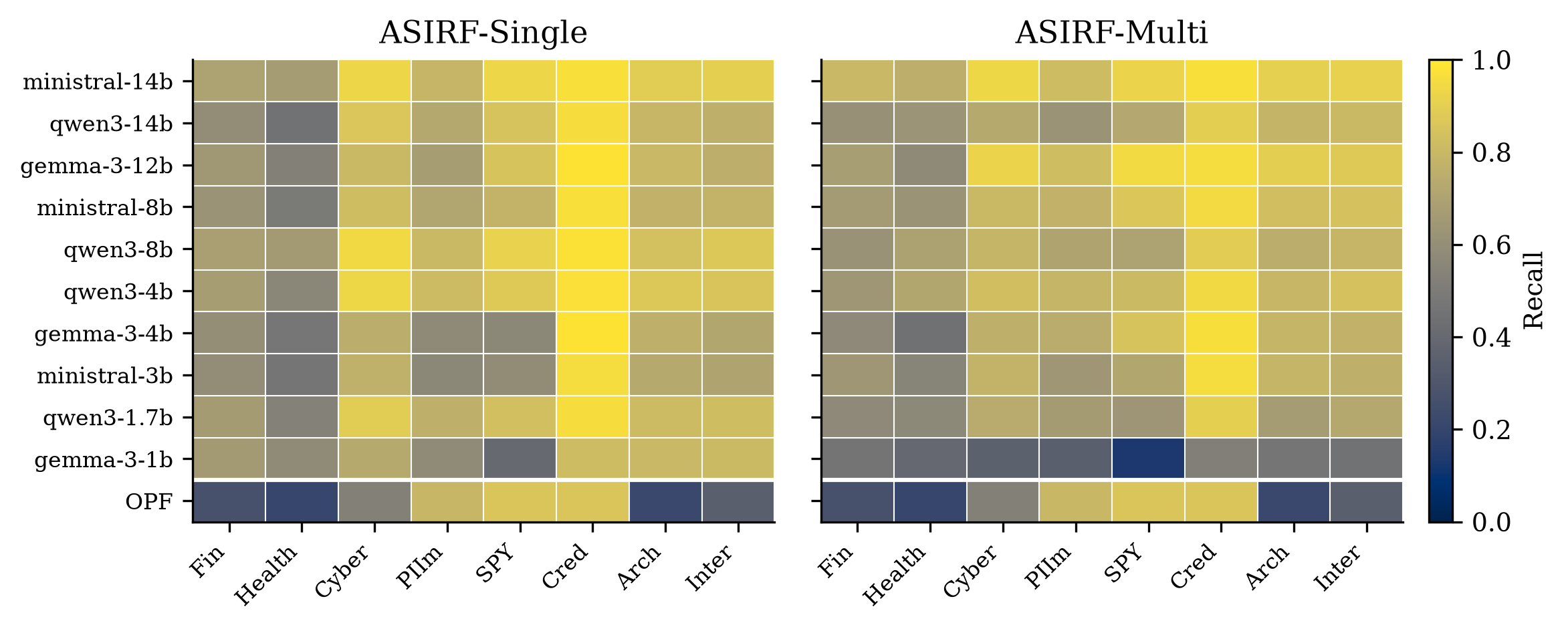}
\caption{Recall, every model plus OPF, across all 8 datasets, Single-Agent and Multi-Agent side by side (same values as Tables 1--2). Datasets: Fin=Finance, Health=Healthcare, Cyber=Cybersecurity, PIIm=pii-masking, SPY=SPY, Cred=CredData, Arch=Archive of Forgotten Futures, Inter=Interstellar Immigration Bureau.}
\label{fig:recall-heatmap}
\end{figure}


\subsubsection{Out-of-distribution generalization}
\label{sec:oodgen}

Every model, under both architectures, exceeds OPF's fixed recall on the two fictional domains.
This holds even for Gemma-3-1B, the weakest model elsewhere (Figure 2).
The reason is not that fictional domains are easier in general.
It is a difference in cause.
Healthcare and finance are domains a trained classifier like OPF could plausibly handle well too, given how abundant public training data is for them.
Categories like "Temporal Clearance Level," by construction, could never have appeared in any training corpus at all.
A knowledge base can be given a definition for either kind of category but a fixed classifier cannot recognize the latter at all without expensive training, which is the clearest evidence for the paper's central claim.

\subsection{Ablation}
\label{sec:ablation}

Removing the Entity Retriever from ASIRF-Multi isolates the contribution of the knowledge base on a scoped grid (three models,
Archive of Forgotten Futures custom dataset), confirming that retrieval contributes to better detection of
sensitive values and improves recall across all three models (Table 3).

\begin{table}[t]
\centering
\caption{Recall on Archive of Forgotten Futures custom dataset, by model and architecture.}
\label{tab:ablation}
\begin{tabular}{lccc}
\toprule
Model & ASIRF-Multi & No-RAG \\
\midrule
ministral-14b & \textbf{0.908} & 0.698 \\
qwen3-14b     & \textbf{0.783} & 0.750 \\
gemma-3-12b   & \textbf{0.902} & 0.877 \\
\bottomrule
\end{tabular}
\end{table}

\section{Conclusion}

ASIRF, an agentic framework that determines sensitivity via an inference-time knowledge base instead of a taxonomy fixed at training time, was evaluated across ten open-weight models and two architectures. Using a small fraction of a trained detector's preparation cost, its recall exceeds OPF's in 85\% of model–domain combinations, including domains outside OPF's training distribution and fictional domains unrecognizable to any fixed classifier. This efficiency, further improvable by strengthening knowledge-base curation, comes with a trade-off. ASIRF's LLM calls cost more at inference than OPF's single lightweight forward pass. Within the harness itself, ASIRF-Multi and ASIRF-Single each suit different domains and model families rather than either winning uniformly, while use of a knowledge base for retrieval improves recall across models. Future work includes improving the knowledge base, such as a web-search fallback for the Entity Retriever to drive new-domain preparation time toward zero.

\bibliographystyle{unsrt}
\bibliography{ref}


\appendix

\section{Appendix}

\subsection{Precision and F1}

\begin{table}[h]
\centering
\scriptsize
\caption{ASIRF-Multi precision, by model and domain.}
\label{tab:appendix-precision-multi}
\resizebox{\linewidth}{!}{%
\begin{tabular}{lcccccccc}
\toprule
& \multicolumn{6}{c}{Real-world domains} & \multicolumn{2}{c}{Fictional domains} \\
\cmidrule(lr){2-7} \cmidrule(lr){8-9}
Model & Finance & Healthcare & Cybersecurity & pii-masking & SPY & CredData & \shortstack{Archive of\\Forgotten Futures} & \shortstack{Interstellar\\Immigration Bureau} \\
\midrule
ministral-14b & 0.508 & 0.602 & 0.690 & 0.882 & 0.494 & 0.710 & 0.627 & 0.605 \\
qwen3-14b     & 0.612 & \textbf{0.750} & \textbf{0.844} & 0.914 & \underline{0.501} & \textbf{0.828} & \textbf{0.785} & \textbf{0.827} \\
gemma-3-12b   & 0.552 & 0.608 & 0.659 & \underline{0.917} & 0.498 & 0.701 & 0.657 & 0.654 \\
ministral-8b  & 0.539 & 0.592 & \underline{0.812} & 0.900 & 0.495 & 0.795 & 0.696 & 0.671 \\
qwen3-8b      & 0.609 & 0.707 & 0.801 & 0.911 & 0.478 & \underline{0.815} & 0.728 & \underline{0.740} \\
qwen3-4b      & 0.597 & 0.664 & 0.766 & 0.900 & 0.493 & 0.804 & 0.684 & 0.683 \\
gemma-3-4b    & \underline{0.615} & 0.658 & 0.788 & 0.916 & 0.482 & 0.735 & \underline{0.761} & 0.739 \\
ministral-3b  & 0.591 & 0.664 & 0.762 & 0.882 & 0.491 & 0.756 & 0.712 & 0.653 \\
qwen3-1.7b    & 0.545 & 0.684 & 0.709 & 0.900 & \textbf{0.546} & 0.779 & 0.612 & 0.596 \\
gemma-3-1b    & 0.365 & 0.496 & 0.417 & 0.787 & 0.278 & 0.630 & 0.453 & 0.409 \\
\midrule
\textbf{OPF}    & \textbf{0.676} & \underline{0.730} & 0.727 & \textbf{0.932} & 0.441 & 0.811 & 0.669 & 0.712 \\
\bottomrule
\end{tabular}%
}
\end{table}

\begin{table}[h]
\centering
\scriptsize
\caption{ASIRF-Single precision, by model and domain.}
\label{tab:appendix-precision-single}
\resizebox{\linewidth}{!}{%
\begin{tabular}{lcccccccc}
\toprule
& \multicolumn{6}{c}{Real-world domains} & \multicolumn{2}{c}{Fictional domains} \\
\cmidrule(lr){2-7} \cmidrule(lr){8-9}
Model & Finance & Healthcare & Cybersecurity & pii-masking & SPY & CredData & \shortstack{Archive of\\Forgotten Futures} & \shortstack{Interstellar\\Immigration Bureau} \\
\midrule
ministral-14b & 0.535 & 0.648 & 0.658 & 0.920 & 0.508 & 0.793 & 0.628 & 0.596 \\
qwen3-14b     & \textbf{0.704} & \textbf{0.803} & 0.791 & \textbf{0.943} & \textbf{0.537} & \underline{0.795} & 0.807 & \textbf{0.814} \\
gemma-3-12b   & 0.628 & 0.746 & 0.831 & 0.903 & 0.501 & 0.731 & 0.791 & 0.741 \\
ministral-8b  & 0.677 & \underline{0.797} & 0.865 & 0.931 & \underline{0.529} & 0.751 & 0.808 & 0.690 \\
qwen3-8b      & 0.694 & 0.743 & 0.807 & 0.931 & 0.520 & 0.738 & 0.752 & 0.726 \\
qwen3-4b      & 0.672 & 0.666 & 0.736 & 0.928 & 0.523 & 0.780 & 0.696 & 0.653 \\
gemma-3-4b    & \underline{0.694} & 0.786 & \textbf{0.871} & 0.907 & 0.485 & 0.768 & \underline{0.808} & 0.729 \\
ministral-3b  & 0.685 & 0.789 & \underline{0.870} & 0.927 & 0.524 & 0.792 & \textbf{0.811} & \underline{0.778} \\
qwen3-1.7b    & 0.519 & 0.609 & 0.660 & 0.921 & 0.502 & 0.757 & 0.616 & 0.599 \\
gemma-3-1b    & 0.338 & 0.486 & 0.491 & 0.880 & 0.474 & 0.689 & 0.453 & 0.480 \\
\midrule
\textbf{OPF}    & 0.676 & 0.730 & 0.727 & \underline{0.941} & 0.441 & \textbf{0.811} & 0.669 & 0.712 \\
\bottomrule
\end{tabular}%
}
\end{table}

\begin{table}[h]
\centering
\scriptsize
\caption{ASIRF-Multi F1, by model and domain.}
\label{tab:appendix-f1-multi}
\resizebox{\linewidth}{!}{%
\begin{tabular}{lcccccccc}
\toprule
& \multicolumn{6}{c}{Real-world domains} & \multicolumn{2}{c}{Fictional domains} \\
\cmidrule(lr){2-7} \cmidrule(lr){8-9}
Model & Finance & Healthcare & Cybersecurity & pii-masking & SPY & CredData & \shortstack{Archive of\\Forgotten Futures} & \shortstack{Interstellar\\Immigration Bureau} \\
\midrule
ministral-14b & \textbf{0.622} & 0.672 & 0.794 & 0.850 & \underline{0.643} & 0.821 & 0.742 & 0.727 \\
qwen3-14b     & 0.611 & 0.684 & 0.784 & 0.743 & 0.593 & 0.862 & \textbf{0.784} & \textbf{0.817} \\
gemma-3-12b   & 0.608 & 0.594 & 0.769 & \textbf{0.869} & \textbf{0.653} & 0.810 & 0.760 & 0.750 \\
ministral-8b  & 0.595 & 0.609 & \textbf{0.809} & 0.834 & 0.630 & \underline{0.866} & 0.758 & 0.749 \\
qwen3-8b      & 0.615 & \textbf{0.700} & 0.794 & 0.795 & 0.568 & 0.851 & 0.741 & \underline{0.765} \\
qwen3-4b      & \underline{0.619} & \underline{0.689} & \underline{0.796} & 0.840 & 0.613 & \textbf{0.868} & 0.735 & 0.755 \\
gemma-3-4b    & 0.596 & 0.533 & 0.776 & 0.824 & 0.617 & 0.835 & \underline{0.773} & 0.755 \\
ministral-3b  & 0.615 & 0.602 & 0.771 & 0.740 & 0.583 & 0.845 & 0.748 & 0.704 \\
qwen3-1.7b    & 0.561 & 0.623 & 0.724 & 0.766 & 0.587 & 0.837 & 0.639 & 0.655 \\
gemma-3-1b    & 0.408 & 0.441 & 0.384 & 0.485 & 0.177 & 0.570 & 0.462 & 0.429 \\
\midrule
\textbf{OPF}    & 0.386 & 0.321 & 0.613 & \underline{0.861} & 0.584 & 0.836 & 0.327 & 0.468 \\
\bottomrule
\end{tabular}%
}
\end{table}

\begin{table}[h]
\centering
\scriptsize
\caption{ASIRF-Single F1, by model and domain.}
\label{tab:appendix-f1-single}
\resizebox{\linewidth}{!}{%
\begin{tabular}{lcccccccc}
\toprule
& \multicolumn{6}{c}{Real-world domains} & \multicolumn{2}{c}{Fictional domains} \\
\cmidrule(lr){2-7} \cmidrule(lr){8-9}
Model & Finance & Healthcare & Cybersecurity & pii-masking & SPY & CredData & \shortstack{Archive of\\Forgotten Futures} & \shortstack{Interstellar\\Immigration Bureau} \\
\midrule
ministral-14b & 0.606 & \underline{0.659} & 0.771 & 0.851 & 0.657 & \textbf{0.874} & 0.737 & 0.719 \\
qwen3-14b     & 0.645 & 0.578 & 0.826 & 0.821 & \underline{0.658} & \underline{0.871} & \textbf{0.800} & \underline{0.789} \\
gemma-3-12b   & 0.636 & 0.618 & 0.818 & 0.772 & 0.631 & 0.838 & \underline{0.796} & 0.750 \\
ministral-8b  & 0.649 & 0.610 & \underline{0.846} & 0.808 & 0.630 & 0.848 & 0.792 & 0.733 \\
qwen3-8b      & \textbf{0.689} & \textbf{0.699} & \textbf{0.870} & \underline{0.863} & \textbf{0.663} & 0.841 & 0.794 & \textbf{0.792} \\
qwen3-4b      & \underline{0.673} & 0.609 & 0.824 & \textbf{0.868} & 0.655 & 0.867 & 0.775 & 0.742 \\
gemma-3-4b    & 0.644 & 0.593 & 0.807 & 0.708 & 0.524 & 0.861 & 0.786 & 0.724 \\
ministral-3b  & 0.637 & 0.589 & 0.815 & 0.704 & 0.555 & 0.867 & 0.769 & 0.740 \\
qwen3-1.7b    & 0.584 & 0.568 & 0.757 & 0.836 & 0.626 & 0.848 & 0.701 & 0.694 \\
gemma-3-1b    & 0.447 & 0.530 & 0.588 & 0.701 & 0.436 & 0.749 & 0.579 & 0.604 \\
\midrule
\textbf{OPF}    & 0.386 & 0.321 & 0.613 & 0.862 & 0.584 & 0.836 & 0.327 & 0.468 \\
\bottomrule
\end{tabular}%
}
\end{table}

\subsection{Compute resources}

All model inference was performed via paid-tier hosted inference APIs across two providers, Amazon Bedrock and Hugging Face (Inference Endpoints/paid inference access, not the free tier), rather than self-hosted GPU compute. Local compute was limited to lightweight orchestration, ChromaDB retrieval, and scoring, none of which required GPU acceleration. 

Per-row processing time varies by dataset, reflecting differences in input length and reasoning complexity rather than a fixed per-call cost. Average processing time was 5 s/row for one representative model (Qwen3-14B, Single-Agent).


\end{document}